\def\year{2018}\relax
\documentclass[letterpaper]{article} 
\usepackage{aaai18}  
\usepackage{times}  
\usepackage{helvet}  
\usepackage{courier}  
\usepackage{url}  
\usepackage{graphicx}  
\usepackage{amssymb}
\usepackage{amsmath}
\usepackage{color}
\usepackage{graphicx}
\graphicspath{ {figs/} }
\usepackage{calc}
\usepackage{blindtext}
\usepackage{enumitem}
\usepackage{scrextend}
\addtokomafont{labelinglabel}{\sffamily}
\usepackage{mwe}
\usepackage{booktabs}
\usepackage{multirow}
\usepackage{siunitx}

\newcommand{\newcite}[1]{\citeauthor{#1} (\citeyear{#1})}

\DeclareMathOperator{\E}{\mathbb{E}\,}
\DeclareMathOperator{\embb}{c}
\DeclareMathOperator{\loss}{\mathcal{L}}

\title{Source-Target Inference Models for Spatial Instruction Understanding}

\author{Hao Tan \and Mohit Bansal\\
Department of Computer Science\\
University of North Carolina at Chapel Hill\\
\texttt{\{haotan, mbansal\}@cs.unc.edu}\\
}

\date{}

\begin{document}

\maketitle

\begin{abstract}
Models that can execute natural language instructions for situated robotic tasks such as assembly and navigation have several useful applications in homes, offices, and remote scenarios.
We study the semantics of spatially-referred configuration and arrangement instructions, based on the challenging Bisk-2016 blank-labeled block dataset. This task involves finding a source block and moving it to the target position (mentioned via a reference block and offset), where the blocks have no names or colors and are just referred to via spatial location features.
We present novel models for the subtasks of source block classification and target position regression, based on joint-loss language and spatial-world representation learning, as well as CNN-based and dual attention models to compute the alignment between the world blocks and the instruction phrases. For target position prediction, we compare two inference approaches: annealed sampling via policy gradient versus expectation inference via supervised regression. Our models achieve the new state-of-the-art on this task, with an improvement of 47\% on source block accuracy and 22\% on target position distance. 

\end{abstract}

\section{Introduction}

\begin{figure}
\centering
\includegraphics[width=0.98\linewidth]{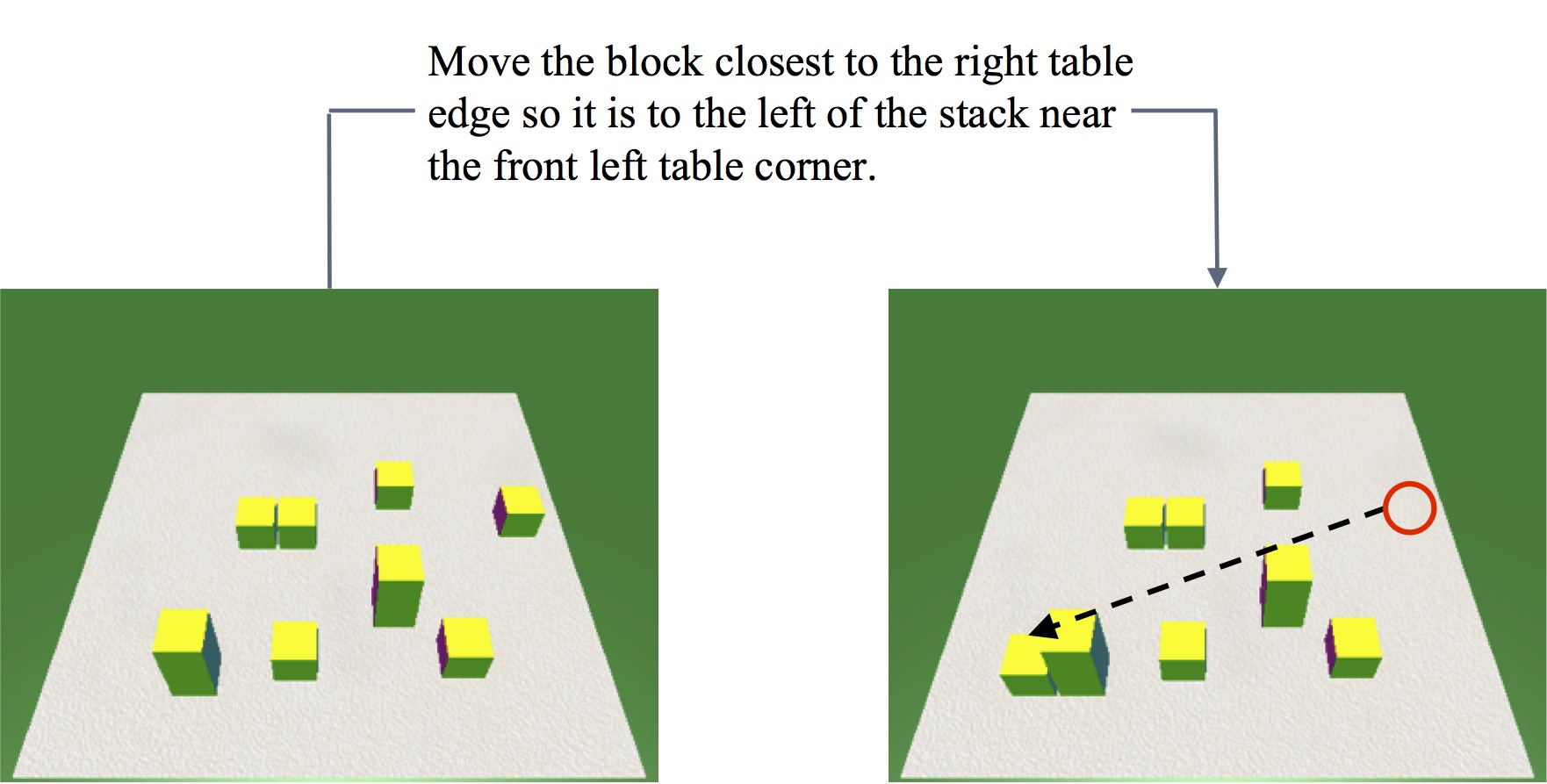}
\caption{An example of the configuration instruction understanding task (based on blank-labeled blocks). Our model is able to correctly predict the source block and the target position in this case. 
}
\label{fig:problem}
\end{figure}

The task of robotic instruction execution involves developing models that can understand the semantics of free-form natural language instructions and execute them as a sequence of actions. Such models have several useful applications in the domain of navigation, manipulation, and assembly, and in the scenarios of homes, offices, warehouses, esp. in remote settings. In this paper, we address the task of executing assembly-style configuration (arrangement) instructions, where the goal is to predict the spatially-referred source block and then move it to the target position, which in turn is referred to in terms of a reference block and an offset to it (again only using spatial features; see Fig.~\ref{fig:problem}). 
Our task is an idealization of the general assembly problem, while still involving similar challenges and features, as well as requiring solutions that can be extended to other robotic instruction problems such as those for navigation and manipulation (e.g., instruction-world alignment with spatial references, sampling with rewards, and joint representation learning across subtasks).

Models that can understand the semantics of block selection and moving instructions (and the involved referring expressions) have been a topic of study since the 1970s, e.g., the SHRDLU system~\cite{winograd72}.
We focus on the recent block-arrangement instructions dataset (its `much more challenging' blank-labeled version) by~\newcite{bisk2016natural}, which is important and challenging because of several reasons. 
First, their instructions are free-form and substantially diverse in language vocabulary and structure, making it hard for a formulaic or pattern-based grammar model to capture the correct semantics. Secondly, the reference to the source block and the target position is solely based on complex spatial-relative information because all the blocks are identical except for their positions (i.e., they have no names, labels, color, etc.). Hence, they involve varying hops of inference and use diverse blocks as their reference (contextual) anchors. Third, the supervision for the target task is only provided directly for the final target position, and not for the intermediate reference block and offset value. Lastly, the dataset size is limited compared to its diversity and complexity.

We propose novel models for this configuration-based instruction understanding dataset and task using joint-subtask-loss representation learning, dual and CNN based attention, and expectation and sampling based inference approaches. 
First, the source, reference, and offset subtask models all share the same sentence and block-location representation parameters (as well as the bilinear attention matrices) so as to learn shared spatial-relative semantics across the subtasks (via optimization of a joint-subtask loss function), given the limited and diverse data. Next, we use advanced and task-suited bilinear attention models based on dual language-to-block relationships and CNN filters, so as to align the different parts of the instruction with appropriate spatial-relative features of the different blocks. Finally, we present two inference and optimization methods to combine the reference and offset values for target prediction: expectation of positions optimized via supervised regression, vs. sampling of an interpretable, single reference block optimized via policy gradient (with effective average-block annealing procedure). 

Empirically, our models achieve substantial improvements over previous work: 47\% on source block selection accuracy and 22\% on target position mean distance.

\section{Related Work}

Starting from the SHRDLU system~\cite{winograd72}, several papers ~\cite{branavan2009reinforcement,howard2014natural,matuszek2014learning} have aimed at building the mapping from natural language instructions to manipulation and assembly style actions on objects. 
To overcome the constraint of using fixed-template instructions, several papers presented mapping based on the induction of semantic grammars~\cite{zettlemoyer2005learning,tellex2011understanding,matuszek2012joint,misra2015environment,paul2016efficient} which allows the instructions to be more complex and more human-like, also addressing spatial concepts of cardinality and ordinality in referring expressions. 
In addition to assembly and manipulation style instruction understanding tasks, a significant amount of work has focused on the navigational instruction understanding task, i.e., mapping or translating a sequence of instructions about navigation in a visual map to a sequence of travel-based actions~\cite{chen2011learning,artzi2013weakly,mei2015listen,andreas2015alignment} . 
 
We focus on end-to-end neural models that can address the joint, multi-step task of source block prediction and then moving it to a target position based on a reference and offset (with supervision provided only for the final target position). To explore such connections between free-form assembly-style configuration instructions and  actions, we use the recent useful dataset created by~\newcite{bisk2016natural}. They created two versions of this dataset: labeled and unlabeled. In the labeled dataset, each block is assigned with a unique pattern, such as a number or a logo. In the unlabeled dataset, the blocks are blank and have no easy names or numbers and hence have to be only referred to by complex spatial-relative features. We focus on this latter, more challenging spatial-semantics dataset and task.

The source block selection model proposed in ~\cite{bisk2016natural} is based on a softmax classifier built on the last hidden state of the instruction LSTM-RNN. The target position prediction is trained using supervised regression.  Their final model is an RNN-based `end-to-end' neural model, which works well on the pattern-labeled dataset but not so well on the blank-labeled dataset (which requires learning the correct position features and world-language alignment). We propose joint-subtask, location-aware, and alignment-learning models for this unlabeled task, and use sampling and expectation based inference to combine the reference and offset values (since supervision is provided only for the final target position); we also learn shared world, language, and attention parameters across the source, reference, and offset tasks.

\begin{figure*}
\centering
\includegraphics[width=0.98\linewidth]{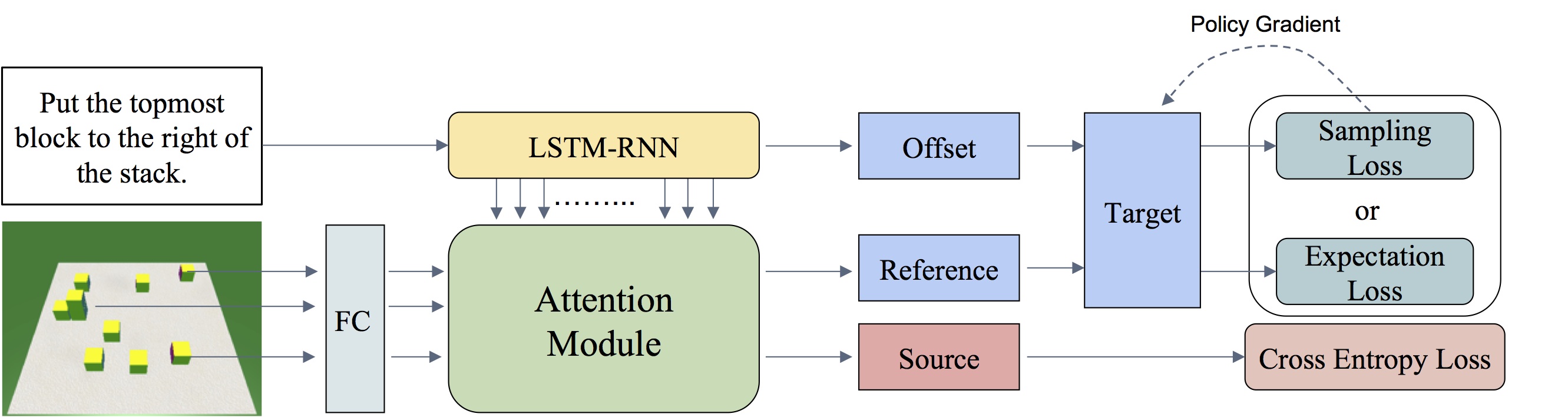}
\caption{Our overall model for the assembly instruction understanding task, showing instruction and world representation learning, language-to-block alignment modules, and source and target (expectation vs. sampling) loss functions.}
\label{fig:model}
\end{figure*}

Related to sampling-based loss and policy gradient optimization,~\newcite{branavan2009reinforcement} adopt policy gradient based reinforcement learning for executing instructions on system troubleshooting and game tutorials.
 There is also recent policy gradient approaches for the tasks of machine translation and image captioning using metric-based rewards~\cite{ranzato2015sequence,xu2015show}. Since the losses of these models are non-differentiable, a policy gradient approach (introduced in~\newcite{williams1992simple}) is used for optimization. 
Most recently,~\newcite{misra2017mapping} extended the pattern-labeled version of the ~\newcite{bisk2016natural} dataset to a new sequential motion planning task based on raw visual simulation input (for intermediate movement steps) fed into a reinforcement learning model. On the other hand, we focus on a different setup, i.e., the original~\newcite{bisk2016natural}  source+target
direct-prediction task and dataset; and we address its more challenging blank-labeled-blocks version, hence only relying on spatial location-based semantics.

\section{Models}
\label{sec:model}

Figure~\ref{fig:model} illustrates our overall model.
Our block-moving task involves two subproblems: source block selection and target position prediction, where the first is a classification problem (among the given set of blocks) and the latter is a regression problem (based on distance error).
The source label and target position are inherently independent of each other and we first present separate models for each. However, the instruction and world representations can benefit from shared knowledge on common spatial terms used to refer to the source, reference, and offset, and hence we also propose a joint model which uses the combined loss function of the source and target tasks to compute gradients that update the shared language LSTM-RNN and the world representation (as well as attention) parameters.

Note that Sec.~\ref{sec:val} presents all the ablation results for various model components and choices discussed in this section.

\subsection{Source Block Selection} \label{source}
We model the source block selection subtask as a classification problem over the finite set of blocks in the world. 
If the input is all the $n$ block positions $B = \{b_1, b_2, \ldots, b_n\}$ and an instruction with $m$ words $I=\{w_1, w_2, \ldots, w_m\}$, the goal is to predict which block does the source part of the instruction refer to. 
The source block selection model consists of two phases: encoding and alignment. For the encoding phase, the instruction and the blocks are encoded into their respective embedding representations. For the alignment phase, an attention module is used to measure the matching between the instruction and each of the block embeddings. Finally, the block which best aligns with the instruction is chosen as the source block.

We first use a standard LSTM-RNN to encode the instruction $I$ into its embedding representation $H = \{h_1, h_2, \ldots, h_m\}$ by the recurrent function:
\begin{equation}
h_t = r\left( h_{t-1}, W_{\textsc{W}} w_t \right)
\end{equation}
where $W_{\textsc{W}}$ is the word embedding matrix layer.
The blocks are encoded into their embeddings $\embb_i$ via a fully-connected layer with a sigmoid activation unit:
\begin{equation}
\embb_i = \sigma(W_\textsc{B} f_{b_i} + a)
\end{equation}
where $f_{b_i}$ is the input feature representation of the $i$th block consisting of its coordinates and its relative distance and stack based features discussed below; $W_\text{B}$ and $a$ are the block weights and bias parameters.

\textbf{Block Features}:
If the block is represented by its absolute coordinate features alone, it will not be aware of its surrounding blocks and relative position on the board, which is important information for understanding spatial instructions, esp. given that the blocks are `blank' (i.e, not labeled with any names or colors), and given the limited size of the dataset. Hence, in addition to the original 3D coordinates, we employ two other simple kinds of relative-position features: (1) The Euclidean distance to each corner and each edge of the board (eight features), (2) A single binary feature indicating whether the block is part of a stack. 

Given these two encoded vectors for the instruction and each block, we next use an attention module to predict the probability of each block being the answer source block (where the source block is represented by the discrete random variable $S$). 
The output of the attention module $\mathcal{A}(\embb_i, H)$ measures the alignment or matching between the $i$th block's embedding $c_i$ and the instruction's embedding $H$. 
Thus, the probability of a block being the source block (given the instruction) is the softmax of the attention value between that block and the instruction sentence.
The source loss function is then the cross-entropy between the conditional distribution $P(S|I)$ and the ground truth distribution $G(B)$ (which is one-hot for the single ground truth block in this task).
\begin{align}
P(S = b_i|I)  \propto \exp(\mathcal{A}(\embb_i, H))\\
\mathcal{L}_\textsc{src}  = -\sum_i\, G(b_i) \log P(S = b_i|I)
\end{align}
This loss function is then summed over for all data instances (instructions) and the total loss is minimized to learn all the source-related weights described above.

Next, we describe the different attention modules that we experimented with.

\subsubsection{Bilinear Attention Modules}
\label{sec:attn}
We use three methods to choose what information from the instruction representation is used to compute the matching with each block representation, each of them employing the bilinear attention form~\cite{luong2015effective}.

\textbf{(1) Last Hidden State}:
The first basic approach simply uses the last hidden vector of the LSTM-RNN $h_m$ and computes the alignment score (with each block vector) using the bilinear form:
\begin{equation}
\mathcal{A}(\embb_i, H) = \embb_i^{\top} W_{\textsc{A}} \, h_m
\end{equation}
where $W_{\textsc{A}}$ is the attention parameter matrix.

\textbf{(2) CNN Filters}:
Instead of only using the last LSTM-RNN hidden vector, we represent the instruction embedding as the concatenation of CNN filters with different kernel sizes, following the idea of sentiment analysis in \cite{kim2014convolutional}. 
We run these different convolutions over the hidden vectors of the LSTM-RNN to compute the outputs of CNN. A max pooling layer is followed to reduce the output sequence of vectors to a single vector $h_{\textsc{CNN}}$, which is then used in the bilinear attention form above.
These CNN filters help capture the key local patterns and hence allow the LSTM to focus on the structure of the sentence. 

\textbf{(3) Dual Attention}:
In our third approach, we develop a novel two-step attention process: word-to-block attention and block-to-instruction attention. 
First, for each block $b_i$, the word-to-block attention part computes the alignment score between each instruction word $w_t$ (represented by its LSTM-RNN hidden state $h_t$) and the block (again using the bilinear form):
\begin{equation}
\text{score}(\embb_i, h_t) = \embb_i^{\top} W_{\textsc{word}} h_t
\end{equation}
Then, the overall context vector $z_i$ for that block $b_i$ is computed as the weighted sum of the LSTM-RNN hidden vectors, with weights based on the softmax of the above word-to-block scores. 
\begin{align}
a_{i,t} & = \frac{\exp \left( \text{score}(\embb_i, h_t) \right) }{\sum_{t'} \exp \left( \text{score}(\embb_i, h_{t'}) \right) } \\
z_i & = \sum_t\, a_{i,t} h_t
\end{align}

Finally, the second stage block-to-instruction attention part computes the block's alignment score with the full instruction as the alignment score between this overall context vector $z_i$ above and the block embedding (via a second bilinear form). 
\begin{equation}
\mathcal{A}(\embb_i, H) = \embb_i^{\top} W_{\textsc{block}} \, z_i
\end{equation}

This dual approach hence allows the model to learn which part (if any) of the sentence refers to each block and get a block-conditioned context vector of the instruction (for each block). Then, in the second stage, each block is compared to its own context vector instead of the global one (as done in the other attention models).

\subsection{Target Position Prediction}

We divide the target position task into the task of finding the reference block and the offset to it, e.g., ``Move the bottom block to the left of the rightmost block''. Here, ``the rightmost block'' is the reference block, while ``to the left'' is the offset.
We next describe the way we model the reference and offset random variables $R$ and $O$ and their training methods, and finally their combination to predict the target position (since the dataset supervision only exists at the final target position level).

\subsubsection{Reference and Offset Distributions}
To model the random variable $R$ (where its output represents the 3D coordinates of the reference block), we assign each block a probability of being the answer reference block. 
Similar to source block selection, the probability of a block being the reference block (given the instruction) is the softmax of the attention value between that block and the instruction sentence, but with a separate attention module $\mathcal{A_R}$ with independent context parameters but shared block-instruction bilinear matrices (see Sec.~\ref{sec:attn}):
\begin{equation}
P(R=b_i | I) \propto \exp(\mathcal{A_R}(\embb_i, H))
\end{equation}

The random variable of the offset $O$ obeys a 3D Gaussian distribution with a fixed variance\footnote{The co-variance matrix is identity times a constant (as a tuned hyperparameter for search space size).} $\Sigma_o$:
\begin{equation}
P(O = o | I) \propto \mathcal{N}(\mu_\text{o},\, \Sigma_\text{o})
\end{equation}
The embedding of the instruction (e.g., the CNN version $h_{\textsc{CNN}}$, described in Sec.~\ref{sec:attn}) is followed by fully-connected layers to generate the (x,y,z) coordinates that represent the center $\mu_o$ of the Gaussian distribution:
\begin{equation}
\mu_\text{o} = W_2 \, \sigma(W_1 h_{\textsc{CNN}} + a_1) + a_2 
\end{equation}

\subsubsection{Inference: Sampling vs Expectation}
\label{sec:inference}
Given the distribution of $R$ and $O$ above, the target position's random variable is $T = R  + O$. We then try two strategies to infer the target position $t$: sampling and expectation. 

The inference by sampling strategy allows us to choose a specific single block as the reference (but makes the loss non-differentiable and hence needs policy gradient; discussed below). The reference block $r$ and the offset $o$ are sampled following their distributions, and are then summed to get the sample $t_\textsc{S}$:
\begin{align}
r & \sim P(R = r) \\
o & \sim P(O = o) \\
t_\textsc{S} & = r + o
\end{align}

For the inference based on expectation, the predicted target position $t_\text{E}$ is the expectation of the random variable $T = R+O$:
\begin{align}
t_\text{E} & = \E\left[R + O\right] \\
  & = \E \left[R\right] + \E\left[O\right] 
\end{align}
Hence, unlike the sampling inference, this uses an expected average value over multiple blocks as the reference.

To be compatible with the two different inference strategies, we use two types of losses: the sampling loss and the expectation loss.
The sampling loss is the expectation (over $R$ and $O$) of the distance between the ground truth $t_\textsc{gt}$ and target random variable $T = R+O$:
\begin{align}
\label{sampling loss}
\mathcal{L}_\textsc{smp}  = \E \left \Vert t_\textsc{gt} - R - O \right \Vert
\end{align}
The expectation loss is the squared distance between the ground truth $t_\textsc{gt}$ and expected position:
\begin{align}
\label{expectation loss}
\mathcal{L}_\textsc{exp} &=\left \Vert t_\textsc{gt} - \E\left[ R+O \right] \right \Vert^2\\
&= \left \Vert t_\textsc{gt} - t_\textsc{E} \right \Vert^2 
\end{align}

We use policy gradient to minimize the non-differentiable sampling loss, following previous work in reward-based reinforcement learning.
The expectation loss is fully differentiable so a supervised regression method is used. We next describe the optimization of these two losses in detail.  

\textbf{Sampling Loss (Policy Gradient)}:
Optimization of the sampling loss needs the gradients of the loss:
\begin{align}
\frac{\partial}{\partial \theta} \mathcal{L}_\textsc{smp} = \frac{\partial}{\partial \theta} \E \left \Vert t_\textsc{gt} - R - O \right \Vert 
\end{align}
A Monte Carlo method is used to approximate the above by sampling a sequence of $K$ reference-offset sample pairs
$\{(r_1, o_1), (r_2, o_2), \ldots, (r_K, o_K)\}$
following the distribution of random variables $R$ and $O$:
\begin{align}
\frac{\partial}{\partial \theta} &\mathcal{L}_\textsc{smp} \approx \frac{1}{K} \sum_i\bigg[\bigg(\frac{\partial \log P(R = r_i)}{\partial \theta} - \nonumber \\
&\frac{1}{2}\frac{\partial (o_i-\mu_\textsc{o})^{\top}\Sigma_\textsc{o}^{-1}(o_i-\mu_\textsc{o})}{\partial \theta}\bigg) \cdot \Vert t_\textsc{gt} - r_i - o_i \Vert \bigg]
\end{align}
The overall gradient is the sum of the above over all data instances. This approach to calculate gradients is equivalent to the REINFORCE algorithm~\cite{williams1992simple}
which has also been used in previous image captioning and classification work~\cite{xu2015show,mnih2014recurrent}.
The negative of the distance $\Vert t_\textsc{gt} - r_i - o_i \Vert$ between the ground truth target $t_\textsc{gt}$ and the target prediction $t_\textsc{s}$ is viewed as the reward.\footnote{To reduce the variance of the estimator, we use the popular `baseline' technique, where a scalar is subtracted from the reward in the updating rule. Specifically, we use the linear-regression baseline~\cite{williams1992simple,ranzato2015sequence}. This was better than the exponential-moving-average~\cite{williams1992simple} and the self-critical~\cite{Rennie2016self} baselines.
} 

\textbf{Annealing Method}:
To solve the instability of the one-block sampling method above (esp. on this challenging small dataset), we slowly anneal the expectation loss $\mathcal{L}_\textsc{EXP}$ to the sampling loss $\mathcal{L}_\textsc{SMP}$ via a sample-averaging intermediate loss $\mathcal{L}_\mathbf{N}$ (related to the annealing method of~\newcite{ranzato2015sequence}).
To build this intermediate loss $\mathcal{L}_\mathbf{N}$, we sample $N$ $($block, offset$)$ pairs $\{(r_1, o_1), \ldots, (r_N, o_N)\}$ following the distribution of the random variables $R$ and $O$, and then we use the average of these pairs as the target prediction (as opposed to using a single reference block and offset sample). The loss $\mathcal{L}_\mathbf{N}$ is then the distance between the ground truth and the prediction.
The motivation is that the expectation loss $\mathcal{L}_{\textsc{EXP}}$ (with L1 norm) and the sampling loss $\mathcal{L}_{\textsc{SMP}}$ are the two limits of this intermediate loss, 
and therefore, starting from the expectation loss, we anneal it to the sampling loss by slowly decreasing the sample size $N$ (details in Sec.~\ref{sec:trainingdetails}).

\textbf{Expectation Loss}:
The loss used here is the sum of the expectation loss $\loss_\textsc{exp} = \left \Vert t_\textsc{gt} - t_\textsc{E} \right \Vert^2$ over the whole dataset. This is similar to the mean squared error (MSE) commonly used for regression. Hence, this is a supervised regression problem which is end-to-end and fully differentiable, and we simply optimize it with a variant of stochastic gradient descent (Adam).

\subsection{Joint Training}
\label{sec:joint}
Although the source and target subproblems represent independent tasks, they still share the same language in the instruction $I$ and the spatial features of the world blocks $\{b_i\}$. For instance, both the source and the reference block are referred to by terms such as `leftmost', `top', etc. and the offset also uses these spatial-directional terms. Hence, we also propose a joint model that learns shared embeddings (across the source and reference tasks) for the words, the LSTM-RNN, and the blocks. Further, the bilinear block-instruction attention matrices are also shared across the source and reference tasks. We optimize the sum of the source loss ($\loss_\textsc{src}$) and the target loss ($\loss_\textsc{tgt} = \loss_\textsc{smp} or \loss_\textsc{exp}$) and compute the gradient of this joint loss to learn the shared parameters.

\section{Experimental Setup}

\begin{table*}[ht!]
\begin{center}
\begin{tabular}{|l|ccc|cc|}
\hline
\multirow{1}{*}{Model Ablation } & \multicolumn{3}{c|}{\textsc{Source}} & \multicolumn{2}{c|}{\textsc{ \ \ \ Target}} \\
  &  \ \ Accuracy \ \  & \ \  Median \ \  &  \ \ Mean  \ \  & \ \  Median \ \  &  \ \ Mean \ \  \\
\hline
Position Features Only				&	50.0\%			&	0.0		&	2.47	&	3.03	&	3.41	\\ 
\hline
Last Hidden State Attention	& 	47.9\%	&	2.28	&	2.88	&	3.15	&	3.48	\\
Dual Attention 	& 	51.4\%	&	0.0		&	2.47	&	3.03	&	3.32	\\
\hline
Annealed Sampling Loss			&	51.3\%		&	0.0	&	2.56	&	3.29	&	3.46	\\
\hline
Non-Joint Training for Source \ \ \ \ \ \ 	&	51.3\%		&	0.0			&	2.53	&	-	&	-	\\
Non-Joint Training for Target	&	-		&	-			& -	&	3.28	&	3.46	\\
\hline
Full Model (Expectation)	&	52.2\%	&	0.0		&	2.44	&	2.91	&	3.23	\\
\hline
\hline
Expectation Model w/ Ensemble	&	54.1\%	&	0.0		&	2.35	&	2.85	&	3.14	\\
Sampling Model w/ Ensemble	&	52.8\%	&	0.0		&	2.41	&	3.09	&	3.25	\\
\hline
\end{tabular}
\end{center}
\caption{Validation results to show ablations of our model components. Our full expectation-based model (third-last row) uses all features (coordinates, relative, stack), CNN attention, and joint training. Each ablation row above that shows the results of changing one component at a time from this full model. Finally, the two last rows represent the final 8-ensemble versions of the full expectation model, as well as the sampling model. (note that lower is better for median and mean distance values)}
\label{table:valid}
\end{table*}

\subsection{Dataset}
We employ the challenging blank-labeled dataset introduced in~\newcite{bisk2016natural}, where each datum includes the natural language instruction, the positions of all the blocks in the world, and the answer source block index plus the coordinates of the final target position (i.e., no supervision exists for the intermediate reference and offset values). 
To collect instructions, they show automatically-rendered images with a source and target choice to MTurkers, and ask them to give unconstrained, free-form instructions that describe the given movement, without being allowed to refer to any name, pattern, or color of the blocks. 
The dataset contains 3573 instructions for 397 image pairs (starting and final configurations). 
We use the standard training/dev/test splits from~\newcite{bisk2016natural}, and use the dev set for all hyperparameter tuning.
 
\subsection{Metrics}

\noindent\textbf{Source Metrics}:
The primary metric for source block classification is the accuracy of correct prediction over the full set of blocks, intuitively because selecting a nearby wrong block causes the full configuration task to fail, no matter how close the wrong block is to the correct source block.
Following previous work, we also report the mean and median of the (Euclidean) distance errors between the predicted block to the answer block coordinates across the dataset (also to be compatible with the target position task). The distance errors are computed in terms of block lengths.

\noindent\textbf{Target Metrics}:
Similar to the source case above, we report the mean and median of the distance errors (computed in terms of block lengths) between the predicted and the ground truth target position coordinates (across the dataset).

\subsection{Training Details}
\label{sec:trainingdetails}
The sentence encoder is an LSTM-RNN with 256-dimensional hidden vectors and word embeddings. The block embedding layer is a 64-dimensional fully-connected layer. We use a generalized (adapted to our approach) Xavier initialization ~\cite{glorot2010understanding} to keep the variance (energy) of each feature map constant across layers, which stabilizes the training process. The Adam optimizer ~\cite{kingma2014adam} is used to update the parameters, and the learning rate is fixed at 0.001. Gradient clipping~\cite{pascanu2013difficulty} is applied to the LSTM parameters to avoid exploding gradients. 
For our annealment-based sampling approach (Sec.~\ref{sec:inference}), we start from the expectation loss, then sample $N=20$ (which approximately matches the expectation loss), and then anneal it down to $1$ (which is same as the one-block sampling loss). To speed up the training process, the initial annealing decay step is 5, which is then reduced to 2, and finally to 1. The final sequence of block samples $N$ is $\{20, 15, 10, 8, 6, 5, 4, 3, 2, 1\}$.

\textbf{Regularization}: To regularize the network, we use weight decay for all trainable variables, and a dropout layer of 0.2 probability is added before and after the LSTM layer. 

\textbf{Data Noising}: To stabilize (or regularize) training on this limited-size dataset, we add two types of noise (only to the training set, not to the validation or test sets): local noise and global noise. The local noise is a 2D Gaussian with a 0.1 (block length) standard deviation, which is added to each block independently. The global noise is a shift of the board, which adds a 2D Gaussian with a 1.0 (block length) standard deviation to the coordinate system. We only add this global noise to the continuous target position variable. Empirically, we found that this small amount of added Gaussian noise has no significant influence on the relationship between the blocks;
but it helps stabilize (or regularize) the training.

\textbf{Pretrained Language Embeddings}: We also tried initializing the instruction's LSTM-RNN with pretrained GloVe word embeddings~\cite{pennington2014glove} but did not see significant improvements, most likely because these embeddings, trained based on unsupervised word context, will not be able to differentiate between identical-context spatial terms like ``left'' and ``right'' in our task. Hence, we allow our embeddings to be trained from scratch directly on the task supervision.

\section{Results and Analysis}

\begin{table*}[ht!]
\begin{center}
\begin{tabular}{|l|ccc|cc|}
\hline
    \multirow{1}{*}{Model} & \multicolumn{3}{c|}{\textsc{Source}} & \multicolumn{2}{c|}{\textsc{ \ \ \ Target}} \\
  &  \ \ Accuracy \ \  & \ \  Median \ \  &  \ \ Mean  \ \  & \ \  Median \ \  &  \ \ Mean \ \  \\
\hline
\hline
End-to-End FFN ~\cite{bisk2016natural} 
& 9.0\% & 3.45 & 3.52 & 3.60 & 3.94 
\\
End-to-End RNN ~\cite{bisk2016natural} \ \ \ 			&	10.0\%	&	3.29	&	3.47	&	3.60	&	3.70	\\
 \hline
 \hline
Our Expectation Model \ \ \ 		&	56.1\% &	0.00	&	2.21	&	2.78	&	3.07	\\
Our Sampling Model \ \ \ 		&	56.3\% &	0.00	&	2.18 	&	3.12	&	3.18	\\
\hline
\hline
Our Expectation Model w/ Ensemble \ \ \ 		&	56.6\%	&	0.00	&	2.12	&	2.65	&	2.91	\\
Our Sampling Model w/ Ensemble\ \ \ 		&	56.8\%&	0.00	&	2.11 	&	2.71	&	2.90	\\
\hline
\end{tabular}
\end{center}
\caption{Final test results of our final sampling and expectation models (w/o and w/ ensemble), compared to the previous state-of-the-art on this dataset.}
\label{table:test_result}
\end{table*}

\begin{figure*}[ht!]
\centering
\includegraphics[width=0.99\linewidth]{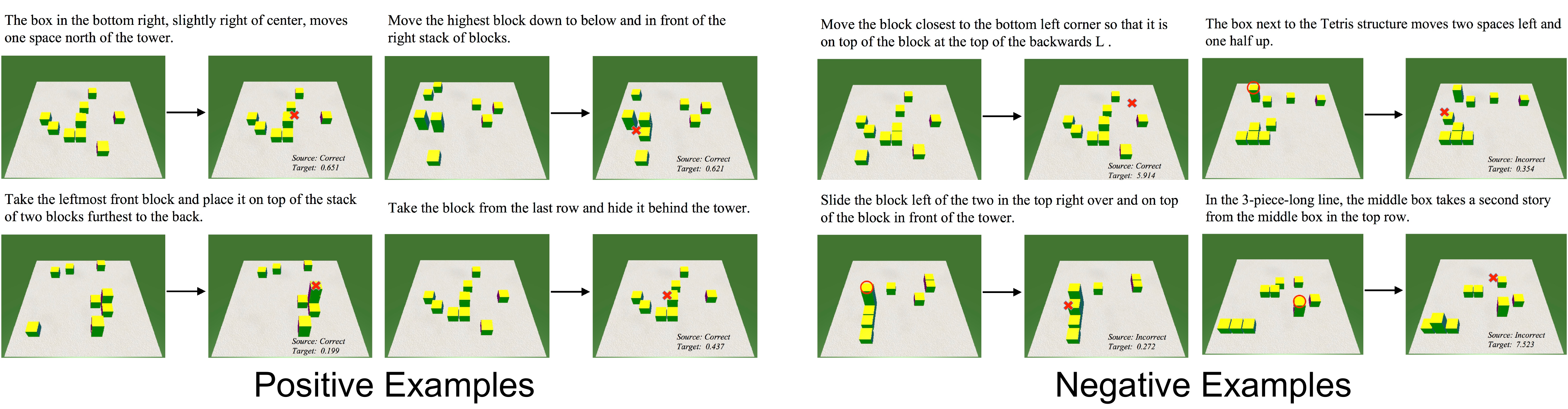}
\caption{Analysis: positive and negative output examples showing interesting instruction scenarios. The first and second image in each pair depict the ground truth movement of the source block to the target position. We report predicted source accuracy and target distance in bottom-right of each second image. We also use a red cross to represent our predicted target position (ground truth target position can be inferred directly from the image difference between the first and second image). Also, for the cases where our model predicted an incorrect source, we represent that wrongly-predicted source block by a red circle.
}
\label{fig:examples}
\end{figure*}

\subsection{Ablation Results}
\label{sec:val}
We first discuss all our ablation results, i.e., the effect of the various model components based on the validation results (for the expectation model). Table~\ref{table:valid} shows our four major model component choices as discussed in Section~\ref{sec:model}: the block-world spatial features, the three attention modules, the sampling vs expectation target loss, and the joint vs non-joint training of the embedding representations.  In Table~\ref{table:valid}, the full expectation-based model (third-last row) represents the model which uses all features (coordinates, relative, stack), CNN attention, and joint training. Each ablation row above this third-last row shows the results on changing one component at a time from this full model. Finally, the last two rows of Table~\ref{table:valid} add an 8-sized ensemble to the full expectation (as well as the sampling) model; and this setting is used for the final test results in Table~\ref{table:test_result}.

\textbf{Feature Selection}:
To show the impact of different representations of the world blocks, we compare the results of using just the coordinate values vs our novel relative and stack-based features (discussed in Section~\ref{source}). 
As shown in Table~\ref{table:valid}, utilizing these new features gives us some decent improvements (2\% in source accuracy and 0.18 in target mean distance).

\textbf{Bilinear Attention Modules}:
For both the source block and reference block selection, we model the distributions by three different bilinear attention modules (discussed in Section~\ref{sec:attn}): bilinear matching between the last hidden state of the instruction and each block, CNN filters on top of the LSTM-RNN vectors, and dual word-to-block and block-to-instruction attention. The comparison among these three attention modules is shown in Table~\ref{table:valid}.
The models using CNN filters or dual attention outperform the one with the last hidden state, on the source and the target tasks. The CNN filter attention model is slightly better than the dual attention model, and hence we use that in the final full model. Note that the dual attention model is similar (within standard deviation) to the CNN attention model in performance. In Sec.~\ref{sec:finaltest}, we also discuss the complementary nature of the CNN and dual attention models, and report their improved combination results.

\textbf{Target Training Methods}:
As shown in Table~\ref{table:valid} (and discussed in Section~\ref{sec:inference}), the model for target prediction is trained with two types of inference methods and optimization loss functions. Using the expectation loss gives us slightly better performance than using the sampling loss (a 0.23 decrease in the validation target mean prediction, and a 0.11 decrease after an ensemble).\footnote{Note that the vanilla, standard sampling approach performed significantly worse than our annealing-based method (described in Sec.~\ref{sec:inference}), achieving a target median of 3.57 and a target mean of 3.82 on the dev set.}
This is likely because the two losses use quite different inference procedures. The sampling inference explicitly chooses (samples) one block as the reference block while the expectation inference calculates the reference by the expected (weighted) sum of several blocks, and both inference choices have their advantages vs. disadvantages (e.g., the sampling method actually allows us to output an interpretable single block as the target reference block, as opposed to the expectation approach), hence we report results for both models.

\textbf{Joint Training}:
The fourth part of Table~\ref{table:valid} compares  joint training vs non-joint training of the world-block and language representations across the source and target tasks (as discussed in Sec.~\ref{sec:joint}). The non-joint training results for the source and the target tasks are worse than the joint training results, showing the advantage of learning shared spatial world and language representations across source and target tasks, via joint loss function optimization. 

Finally, the last row of Table~\ref{table:valid} shows the added effects of an 8-sized standard ensemble approach.

\subsection{Final Test Results}
\label{sec:finaltest}
Next, in Table~\ref{table:test_result}, we present the test-set results for our two inference approaches (expectation and sampling), using the final model choices based on the ablation studies (i.e., all features, CNN attention, joint training), without and with ensemble.
Both inference models achieve strong improvements over the previous best work on this dataset from~\newcite{bisk2016natural}, who employ three neural models for this task. We compare to their final best model, the RNN-based `end-to-end' neural model (as well as their second-best feed-forward network FFN model).
Our model achieves 47\% improvement in source task accuracy, and 22\% (0.8 block length) reduction in target distance mean.
Moreover, the results are quite stable for both inference models: the \textbf{standard deviation} based on 8 runs is around 1\% on source accuracy and 0.05 block length on target mean.

\textbf{Complementarity of Attention Models}:
We found that our two attention models (CNN and dual) are complementary in nature, achieving a source accuracy of \textbf{57.70\%} when combining the ensemble models of CNN and dual attention (for the expectation case), i.e., an improvement of 1.1\% over the CNN model's 56.6\% in Table~\ref{table:test_result}.  
Further experiments in this direction (as well as the complementarity of the sampling and expectation inference approaches) is future work.

\subsection{Analysis}
Figure~\ref{fig:examples} shows several positive and negative examples of the output of our full model. We can correctly understand the semantics in complex source and target descriptions such as `bottom right, slightly right of center' and `place it on top of the stack of two blocks furthest to the back'. In the negative examples, we show complex cases that our model cannot handle correctly, mostly due to special scenarios and phrases that it hasn't seen before in the diverse but small dataset. Examples of this include instructions mentioning shape-based block patterns such as `backwards L', `Tetris structure', and complex count-based patterns such as `3-piece-long line'.

\section{Conclusion}
We presented sampling and expectation based models for source and target prediction in configurational robotic instructions (on a challenging blank-labeled blocks dataset). Our models also use spatial-relative features, CNN and dual attention models, and joint-subtask-loss training of world and language representations, achieving substantial improvements over previous work on all metrics.

\section*{Acknowledgments}
We thank the anonymous reviewers for their helpful
comments. This work was partially supported by a
Google Faculty Research Award, a Bloomberg Data Science Research Grant, an IBM Faculty
Award, and Nvidia GPU awards.

\bibliography{aaai}
\bibliographystyle{aaai}

\end{document}